\documentclass[10pt,twocolumn,letterpaper]{article}

\usepackage[pagenumbers]{cvpr} 
\usepackage{times}
\usepackage{epsfig}
\usepackage{graphicx}
\usepackage{amsmath}
\usepackage{amssymb}
\usepackage{booktabs}
\usepackage{xcolor}
\usepackage{overpic}
\usepackage{multirow,tabularx}
\usepackage{gensymb}
\usepackage{pifont}
\usepackage{adjustbox}
\usepackage{array}
\usepackage{nicematrix}
\usepackage[usestackEOL]{stackengine}
\usepackage{xpatch}
\usepackage[font=small,skip=5pt,belowskip=-9pt]{caption}

\usepackage[safe]{tipa}
\usepackage[utf8]{inputenc}
\usepackage{newunicodechar}
\newunicodechar{ˈ}{\textprimstress}
\newunicodechar{ʃ}{\textesh}
\newunicodechar{ə}{\textschwa}
\newunicodechar{ʊ}{\textupsilon}
\newunicodechar{ɪ}{\textsci}

\definecolor{cvprblue}{rgb}{0.21,0.49,0.74}
\usepackage[pagebackref,breaklinks,colorlinks,citecolor=cvprblue]{hyperref}

\usepackage[capitalize]{cleveref}
\crefname{section}{Sec.}{Secs.}
\Crefname{section}{Section}{Sections}
\Crefname{table}{Table}{Tables}
\crefname{table}{Tab.}{Tabs.}

\def\paperID{61} %
\def\confName{3DV\xspace}
\def\confYear{2024\xspace}

\renewcommand{\paragraph}[1]{\vspace{0.5mm}\noindent\textbf{#1}\:}

\definecolor{limegreen}{HTML}{badc58}
\definecolor{myyellow}{HTML}{f6e58d}

\renewcommand{\vec}[1]{\boldsymbol{#1}}
\newcommand{\mat}[1]{\mathbf{#1}}
\newcommand{\set}[1]{\mathcal{#1}}
\newcommand{\R}{\mathbb{R}}

\newcommand{\loss}{\mathcal{L}}
\newcommand{\lossweight}{\lambda}

\newcommand{\pose}{\vec{\theta}}
\newcommand{\shape}{\vec{\beta}}
\newcommand{\expression}{\vec{\epsilon}}

\newcommand{\id}{\vec{\iota}_g}
\newcommand{\appid}{\vec{\iota}_a}

\newcommand{\point}{\vec{x}}

\newcommand{\imghum}{M}
\newcommand{\dist}{d}
\newcommand{\imghumdist}{\dist_\imghum}
\newcommand{\residualdist}{\Delta\dist}
\newcommand{\col}{\vec{c}}
\newcommand{\semantics}{\vec{s}}
\newcommand{\sd}{(\imghumdist, \semantics)}

\newcommand{\geo}{\mathcal{S}}
\newcommand{\weights}{\vec{\phi}}

\newcommand{\globalfn}{f}
\newcommand{\geofn}{g}
\newcommand{\geofeat}{\vec{g}}
\newcommand{\colfn}{a}
\newcommand{\featureset}{\set{Z}}

\newcommand{\mesh}{\mathcal{M}}
\newcommand{\normal}{\vec{n}}
\newcommand{\gtnormal}{\bar{\normal}}
\newcommand{\gtcol}{\bar{\col}}
\newcommand{\pointson}{\set{O}}
\newcommand{\pointsoff}{\set{F}}
\newcommand{\patch}{\set{P}}
\newcommand{\sigmoid}{\psi}
\newcommand{\sharpness}{k}
\newcommand{\signlabel}{l}
\newcommand{\image}{\mat{I}}
\newcommand{\landmark}{\vec{k}}
\newcommand{\landmarks}{\set{K}}

\def\ia{\emph{i.a}\onedot} 
\newcommand{\cmark}{\ding{51}}%
\newcommand{\xmark}{\ding{55}}%
\newcolumntype{R}[2]{%
    >{\adjustbox{angle=#1,lap=\width-(#2)}\bgroup}%
    l%
    <{\egroup}%
}
\newcommand*\rot{\multicolumn{1}{R{35}{1em}}}

\begin{document}

\setlength{\abovedisplayskip}{3.5pt}
\setlength{\belowdisplayskip}{3pt}

\newcommand\blfootnote[1]{%
  \begingroup
  \renewcommand\thefootnote{}\footnote{#1}%
  \addtocounter{footnote}{-1}%
  \endgroup
}

\title{PhoMoH: Implicit Photorealistic 3D Models of Human Heads}

\author{Mihai Zanfir\textsuperscript{*} \and Thiemo Alldieck\textsuperscript{*} \and Cristian Sminchisescu}

\makeatletter
\let\@oldmaketitle\@maketitle%

\renewcommand{\@maketitle}{
    \@oldmaketitle%
    \centering
    \vspace{-7mm}
    \normalsize{Google Research}
    \vspace{5mm}
}

\maketitle

\begin{abstract}
  \vspace{-2mm}
  We present PhoMoH \emph{\textipa{[ˈfoʊ.moʊ]}}, a neural network methodology to construct generative models of photo-realistic 3D geometry and appearance of human heads including hair, beards, an oral cavity, and clothing.
  In contrast to prior work, PhoMoH models the human head using neural fields, thus supporting complex topology.
  Instead of learning a head model from scratch, we propose to augment an existing expressive head model with new features.
  Concretely, we learn a highly detailed geometry network layered on top of a mid-resolution head model together with a detailed, local geometry-aware, and disentangled color field.
  Our proposed architecture allows us to learn photo-realistic human head models from relatively little data.
  The learned generative geometry and appearance networks can be sampled individually and enable the creation of diverse and realistic human heads.
  Extensive experiments validate our method qualitatively and across different metrics.
  \blfootnote{\textsuperscript{*} The first two authors contributed equally.}
\end{abstract}

\vspace{-4mm}
\section{Introduction}
\vspace{-1mm}
\label{sec:intro}
The last two decades have witnessed significant efforts in modeling the human face and head, and considerable progress has been made \cite{egger20203d}. Statistical models capturing a wide range of shapes, facial expressions, and sometimes appearances have been created~\cite{blanz1999morphable,paysan20093d,FLAME:SiggraphAsia2017,xu2020ghum} and successfully deployed to a wide range of perception and synthesis tasks~\cite{tewari2020stylerig,thies2020neural,feng2021learning}. %
With the advent of deep learning, face and head models have more recently been used for synthetic data generation~\cite{wood2021fake}.
For this new use case, the generative properties of modern statistical models help create large corpora of training data that go well beyond the real data that has originally been used to train the models, in terms of human diversity of expression, head pose, and appearance. 
However, the current generation of models is still far from photorealistic -- a shortcoming not only for synthesis, but also for inference.
Additionally, most models capture only the face, or the head without hair, often lack appearance, and are of only medium spatial resolution.
To alleviate such limitations, some methods combine statistical models with additional data such as textures, normal maps, (facial) hair, and accessories~\cite{wood2021fake}.
However, this approach usually requires significant manual work by artists. Moreover, the models are not completely learned from data, and are thus hard to scale.
When aiming for a diverse holistic model of the human head, the majority of current models additionally face the technical barrier of relying on a template mesh with fixed topology.
\begin{figure}
    \centering
    \includegraphics[width=\linewidth]{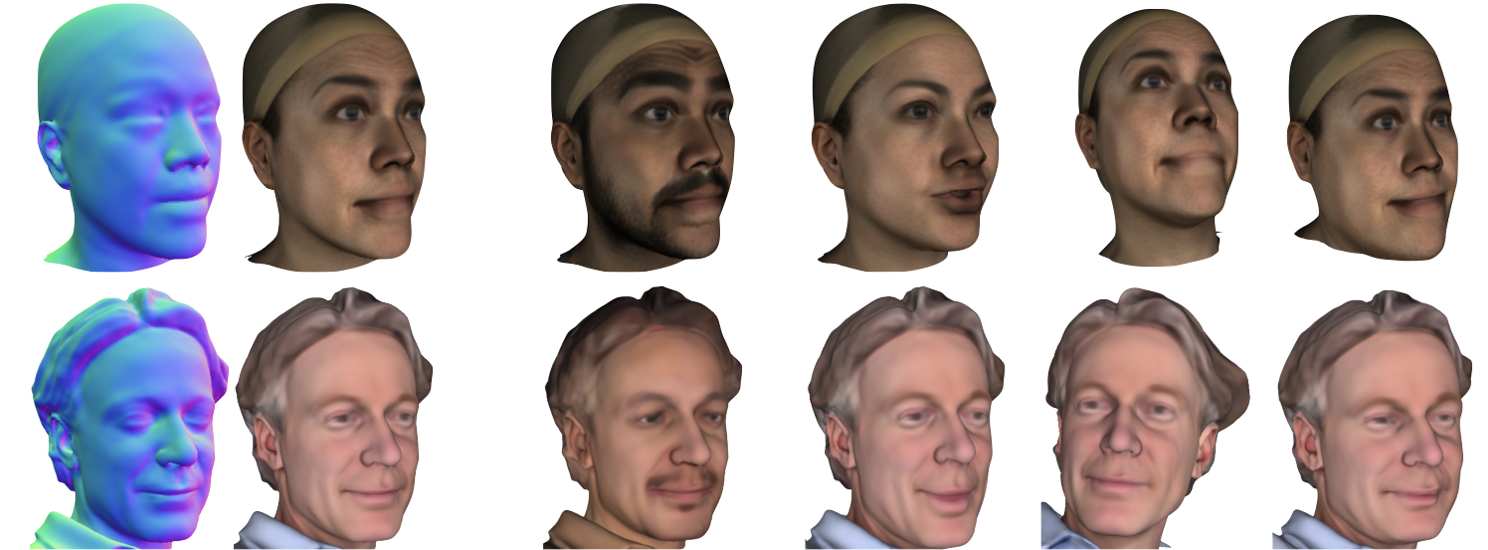}
    \caption{Our implicit generative 3D head model PhoMoH enables disentangled control over a head's geometry, pose and shape, as well as appearance and facial expressions. On the first row, we show a random sample altered using all factors. On the second row, we augment a scan that has been embedded in our model.}
    \label{fig:teaser}
    \vspace{-3mm}
\end{figure}
In practice this means, that \emph{by design}, in the absence of additional components, the models are not easily capable of capturing (facial) hair, clothing, or head shapes which deviate from a generic template.
In this work, we propose to use implicit function networks to go beyond such limitations.
Implicit function networks~\cite{i_OccNet19, i_IMGAN19, i_DeepSDF}, an instance of neural fields \cite{xie2022neural}, model a shape by a decision boundary, or level-set, of a function over points in space.
Consequently, no predefined explicit shape template is used, and the networks can represent varying topology.
This property is particularly useful for modeling the above mentioned desired features, such as hair or accessories, which are missing in the majority of state-of-the-art models.
On the other hand, the current models do have desirable properties, as \eg well-covered face geometry or expression spaces.
To this end, we propose to re-use an existing model and augment it with missing features, instead of going through the laborious and potentially error-prone process of learning a model from scratch.
This way we ensure that our model is on par with state-of-the-art in terms of expressiveness and is capable of additional coverage.
Concretely, we build upon the head component of imGHUM~\cite{alldieck2020imghum} a recent implicit, Signed Distance Function (SDF)-based version of the statistical body model GHUM~\cite{xu2020ghum}.
GHUM and imGHUM can be controlled by a set of generative latent representations for head pose, head shape, and facial expressions, respectively.
Aiming towards photorealism, we augment imGHUM with an additional layer that captures missing factors.
Our joint model, PhoMoH, introduces control over high-resolution geometry including wrinkles and (facial) hair, which we refer to as \emph{identity}.
Additionally, PhoMoH benefits from a detailed, local geometry-aware, and semantically coherent \emph{appearance} model trained using a novel perceptual loss.

To enable PhoMoH's properties, we further introduce a number of \emph{technical contributions}.
\emph{First}, we propose a dual Variational Autodecoder architecture and a training procedure to produce disentangled latent representations for identity and appearance with an imposed Gaussian prior. This enforces compactness and enables latent sampling, interpolation, and optimization.
\emph{Second}, we propose a novel perceptual patch loss with importance sampling to improve the visual
fidelity of our results. 
\emph{Finally}, we introduce a local geometry feature, trained without explicit supervision, thus enabling interesting applications beyond its original use-case of semantically coherent surface coloring.
In summary, we propose PhoMoH, an implicit, photorealistic head model, with control over facial expression, head pose, low-dimensional head shape, and high-resolution geometric identity and appearance detail (see \cref{fig:teaser}).

\section{Related Work}
\vspace{-1mm}
\label{sec:related}

\begin{figure*}
    \centering
    \vspace{-3mm}
    \begin{overpic}[width=0.95\linewidth]{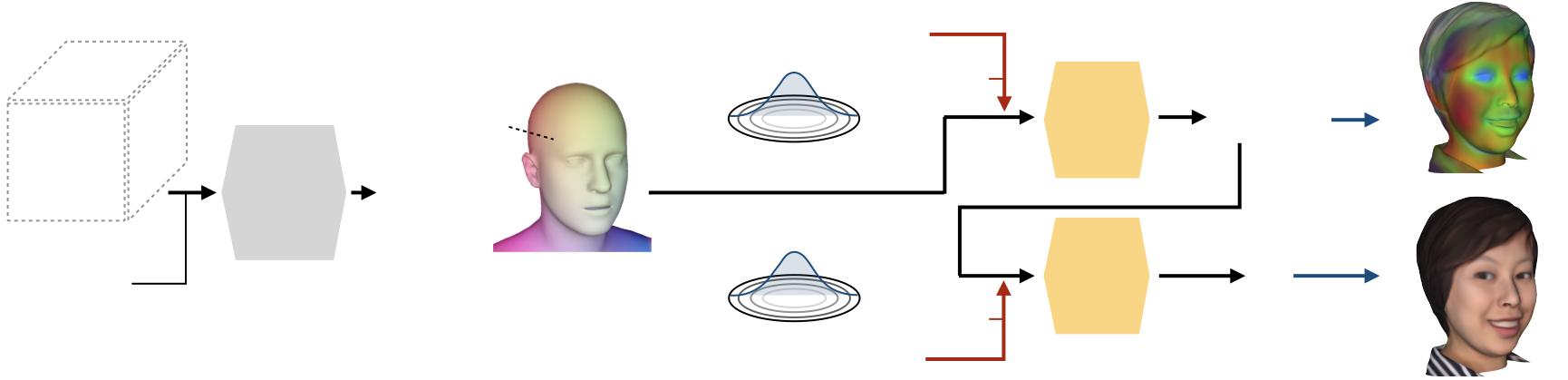} %
    {%
        \put(2, 14.5){\tiny$\bigotimes$}
        \put(3.6,13.4){\small$\point$}
        \put(14.8,11.5){\footnotesize imGHUM}
        \put(1.4,5.5){\small$(\pose,\shape,\expression)$}
        \put(25,11.5){$(\imghumdist, \semantics)$}
        \put(31, 16){\tiny$\bigotimes$}
        \put(45, 21.5){$\id \sim \mathcal{N} (0,\,\mathbf{I})$}
        \put(45, 0.9){$\appid \sim \mathcal{N} (0,\,\mathbf{I})$}
        \put(58.3, 3.5){\small$(\pose,\expression)$}
        \put(58.3, 18.7){\small$(\pose,\expression)$}
        \put(69, 16.1){\LARGE$\geofn$}
        \put(69, 5.8){\LARGE$\colfn$}
        \put(77.8, 16.2){$(\geofeat, \Delta\dist)$}
        \put(80.3, 6.2){$\col$}
    }
    \end{overpic}%
    \vspace{-1mm}
    \caption{Overview of our method. We use imGHUM to canonicalize points in space $\point$ into the pose, shape, and expression-zero-space $\sd$. The geometry network $\geofn$ computes signed distance updates $\residualdist$ to the distance $\imghumdist$ returned by imGHUM. $\geofn$ addionally returns a geometric feature $\geofeat$ that is input to the appearance network $\colfn$ returning a color $\col$ for $\point$. We sample identities 
    by using learned normal distributed latent codes for geometry $\id$ and appearance $\appid$, respectively (red arrows). Using Marching Cubes (blue arrows), we can extract meshes from $\geofn$ and $\colfn$.}
    \label{fig:method}

\end{figure*}

Following the ground-breaking work of Blanz and Vetter \cite{blanz1999morphable}, the realism and the generation capabilities of 3D morphable face and head models (3DMMs) have steadily increased \cite{egger20203d}.
Two distinct lines of works in this space exist, based on (1) explicit mesh-based geometry, and (2) neural implicit surface-based modeling.
The mesh-based representation is by far the most popular in this space, with the implicit one only recently coming into spotlight.

\paragraph{Explicit Models.} A plethora of \emph{face models} have been developed and proposed over the years, all having their roots in the seminal work \cite{blanz1999morphable}, where a low dimensional space is learned for both the appearance and the geometry of the face by using PCA. Subsequent work \cite{paysan20093d, kemelmacher2013internet, Booth_2017_CVPR, gerig2018morphable, booth20163d, booth2018large, cao2013facewarehouse, abrevaya2018multilinear, tran2018nonlinear,feng2021learning,wang2022faceverse}  have expanded on this approach by using better quality or more accessible training data, better registration techniques, or producing more expressive models.
More related are \emph{head models}, that also have seen a rapid development over the years. One of the earliest models, FLAME \cite{FLAME:SiggraphAsia2017}, was trained on 3D head scans, and has decoupled geometric linear latent spaces for shape, pose, and facial expressions. %
\cite{dai2020statistical,dai20173d} introduce a 3D morphable linear model for shape and texture of the full human head. \cite{ploumpis2019combining,ploumpis2020towards} expand on this work to build a more detailed model, with PCA-based controls for the face, cranium, ears and eyes, and basic modeling of the teeth, tongue and inner mouth cavity, as well as an image-based texture completion pipeline.
There are also a number of full body statistical models that support the independent control of the head shape, pose, and expression such as SMPL-X \cite{pavlakos2019expressive}, Frank/Adam \cite{joo2018total}, and GHUM \cite{xu2020ghum}.
For the latter, an implicit formulation, named imGHUM~\cite{alldieck2020imghum}, was presented as well, which we build upon in this work. It is important to note that none of these models are able to represent geometry and/or color for hair or accessories.

\paragraph{Implicit Models} for 3D reconstruction and generation of diverse objects have been introduced in \cite{i_OccNet19, i_IMGAN19, i_DeepSDF}.
In \cite{i_DeepSDF}, a generative auto-decoder based continuous SDF representation is proposed.
Different from an auto-encoder, whose latent code is produced by an encoder, an auto-decoder directly accepts a latent vector as input.
A random latent vector is assigned to each data point in the beginning of training, and the latent vectors are optimized together with the decoder weights.
Implicit models have been quickly adopted for 3D human full body reconstruction \cite{chibane20ifnet,gropp2020igr,saito2020pifuhd,he2021arch++,alldieck2022phorhum} and modeling \cite{deng2019nasa, LEAP:CVPR:2021,Mihajlovic:CVPR:2022}.
In the context of modeling 3D faces and heads, a few approaches have been explored in recent work and are the most related to ours. i3DMM \cite{yenamandra2021i3dmm} is a deep implicit 3D morphable model of the full head with learned latent spaces for geometry and color. The full geometric code is decomposed into local codes controlling identity, expression and hairstyle, while the full color code is further decomposed into identity and hairstyle. Their architectures supports the disentanglement between different latent factors, but does not allow the color field to be informed by semantic signals associated with the underlying geometric field. This limits the plausibility of samples, as for example ear colors could appear splattered onto hair geometry. One additional drawback is the fact that geometry is not metric, limiting some of the practical applications. 
NPHM \cite{giebenhain2022nphm} is a recent implicit head model, represented using an ensemble of local neural fields. However, NPHM only models the geometry and heavily relies on non-rigid registration to a template mesh.
In contrast, our method bypasses this complex pre-processing step inherited from mesh-based models and additionally models disentangled appearance.
ImFace \cite{zheng2022imface} learns a nonlinear implicit representations of only the face geometry, with control over the identity and expression. Related is also LISA \cite{corona2022lisa}, an implicit generative hand model with disentangled shape, pose and appearance parameters.
HeadNerf \cite{hong2021headnerf} is a neural radiance field (NeRF)-based \cite{mildenhall2020nerf} head model that integrates NeRF with a parametric representation of the human head. %
Like in most NeRF-based approaches, plausible geometry is not guaranteed. In recent work, MoRF \cite{wang2022morf} learns highly detailed radiance fields of human heads and allows morphing between identities.
In contrast, our method produces actual 3D geometry (instead of only images) and allows for sampling diverse 3D heads.

\paragraph{Building personalized 3D face or head avatars} from a few images \cite{cao2016real,ramon2021h3d, burkov2022multineus, Khakhulin2022ROME}, or monocular video \cite{garrido2013reconstructing, grassal2021neural, gafni2021nerface,zheng2022imavatar,phonescan} of an individual is a \emph{related but complementary line of work}, which often relies on a statistical model for reconstruction, or as basic representation to further personalize. For instance, H3D-Net \cite{ramon2021h3d} presents an implicit model for 3D head reconstruction from a few images, with a coarse geometric prior trained on an rich dataset of 3D head scans. NeRF-based models have been also used for personalization, either to represent the full person \cite{peng2021neural,xu2021hnerf,liu2021neural,kwon2021neural,weng2022humannerf} or by focusing on the head \cite{gafni2021nerface,park2021nerfies,athar2022rignerf}. In contrast to our work, all these methods estimate personalized 3D heads or faces of individuals, do not build a statistical model, and do not allow for the generation of novel identities. Our model could also be used in a personalization process, either as a statistical prior or as a base representation to further refine. 
Also worth mentioning are 3D-aware GAN models \cite{chan2022efficient} that produce realistic synthesized images with inaccurate 3D frontal geometry and have limited controllability.

\section{Background}
\vspace{-1mm}
\label{sec:background}
Our model builds on top of imGHUM~\cite{alldieck2020imghum}, an implicit model of human pose, shape, and facial expression.
imGHUM is an implicit SDF-based model compatible with the explicit mesh-based model GHUM~\cite{xu2020ghum}, meaning it shares the same low-dimensional parameterization.
GHUM and imGHUM model the human body without hair or high-resolution details such as fine wrinkles and also lack an appearance model.
imGHUM is composed of four sub-models for critical parts of the human body -- please see the original paper for details.
We use the sub-model for the head as the base representation for PhoMoH and simply refer to it as imGHUM hereafter.
imGHUM is a function over a point space $\point$, the head pose $\pose$, the low-dimensional head shape $\shape$, and the facial expression $\expression$, returning the signed distance $\imghumdist$ at $\point$ \wrt the model's surface.
Further, imGHUM returns $\semantics \in \R^{3}$, the nearest neighbor of $\point$ on the surface represented as coordinates on a reference mesh.
Inspired by \cite{xu2021hnerf}, we use imGHUM as a function mapping $\point$ into a reference frame defined through $\sd$ and learn our model in this reference frame.
We illustrate the mapping into $\sd$-space by looking at a concrete example: While the tip of the nose will be located at different points in space $\point$ depending on the head pose or shape, it will always be mapped to the same $\sd$ code.
This is similar to mesh-based modeling, where the tip of the nose will always be modeled by the same vertex.
By building our model in the $\sd$-space, we exploit two properties from imGHUM: (1) our model does not need to learn the range of human shapes, poses, and expression, since these are already well covered by imGHUM and (2) imGHUM serves as an implicit canonicalization function allowing us to learn from pairs of imGHUM codes and unstructured 3D data.

\section{Method}
\vspace{-1mm}
\label{sec:method}

We use template-free neural fields to model both geometry and appearance for human heads.
Our model is controlled by disentangled factors $\featureset = (\pose,\shape,\expression,\id,\appid)$, including head pose $\pose$, low-dimensional head shape $\shape$, facial expression $\expression$, geometric identity $\id$, and appearance  $\appid$.
Concretely, we model the human head as a function over a point in space $\point$ returning the signed distance $\dist$ at $\point$ \wrt the head surface, together with its color $\col$.
The surface $\geo$ of the head is implicitly defined by the zero-level-set of the distance-field modeled by the neural-network $\globalfn$ parameterized by network weights $\weights$
\begin{equation}
    \geo_{\weights}(\featureset) = \Big\{ \point \in \mathbb{R}^3 \: | \: \globalfn\big(\point, \featureset; \weights \big) = (0, \col) \Big\}.
\end{equation}
At test-time we want to be able to sample different appearances for a given geometry.
Thus, we decompose $\globalfn$ into the two multilayer perceptrons (MLPs) $\geofn$ modeling the geometry, and $\colfn$ modeling the appearance.
However, we argue that appearance should follow the semantics defined by  geometry.
For example, skin color should be only placed on skin structure, whereas hair color should only be returned for regions corresponding to hair.
This means that while appearance should be disentangled from specific geometric identities, it should still be informed by the semantic structure defined by $\geofn$.
To this end, we condition $\colfn$ on a low-dimensional surface descriptor $\geofeat$ instead of a point $\point$.
$\geofeat$ is an additional output of $\geofn$ and is learned without explicit supervision. 
This is in a similar spirit as the global geometry feature vector in IDR~\cite{yariv2020multiview}. While IDR performs per object optimization and shows appearance transfer only for uniform materials, our surface descriptor $\geofeat$ allows for complex and semantically meaningful coloring of a whole object class. Moreover, we also apply layer normalization ($LN$) to $\geofeat$ before passing it to the appearance network. In practice this is helpful to prevent information leakage from the geometric identity code into the appearance network and thus helps to disentangle the model parameters, as we will show. 
Besides being a local descriptor for appearance, $\geofeat$ is additionally useful for various applications, see \cref{sec:applications}.
Finally, we learn our model in unposed, neutral expression, and mean-shape space, which we achieve by canonicalizing $\point$ to $\sd$ using imGHUM $\imghum$. The distance returned by imGHUM $\imghumdist$ is a good approximation of the true surface distance $\dist$, thus we let $\geofn$ only find a residual distance $\residualdist$.
The full model $\globalfn$ reads as
\begin{equation}
    \globalfn(\point, \featureset; \weights) = (\imghumdist + \residualdist, \col) = (\dist, \col),
\end{equation}
with
\begin{equation}
    \colfn(LN(\geofeat), \appid, \pose, \expression; \weights_\colfn) = \col,
\end{equation}
\begin{equation}
    \geofn(\imghumdist, \semantics, \id, \pose, \expression; \weights_\geofn) = (\residualdist, \geofeat),
\end{equation}
\begin{equation}
    \imghum(\point, \pose, \shape, \expression) = \sd.
\end{equation}
We additionally condition $\geofn$ and $\colfn$ on pose $\pose$ and expression $\expression$ to capture pose and expression dependent geometry and appearance variations not modeled by imGHUM.
See \cref{fig:method} for an overview. 
In the following we will use $\dist_{\point}$ as shorthand notation for the distance, $\geofeat_{\point}$ for the geometric feature and $\col_{\point}$ for the color returned by $\globalfn$ (or $\geofn$, respectively) at point $\point$, and drop the dependence on $\featureset$ and $\weights$ for clarity.

\begin{figure*}
    \centering
    \includegraphics[width=\linewidth]{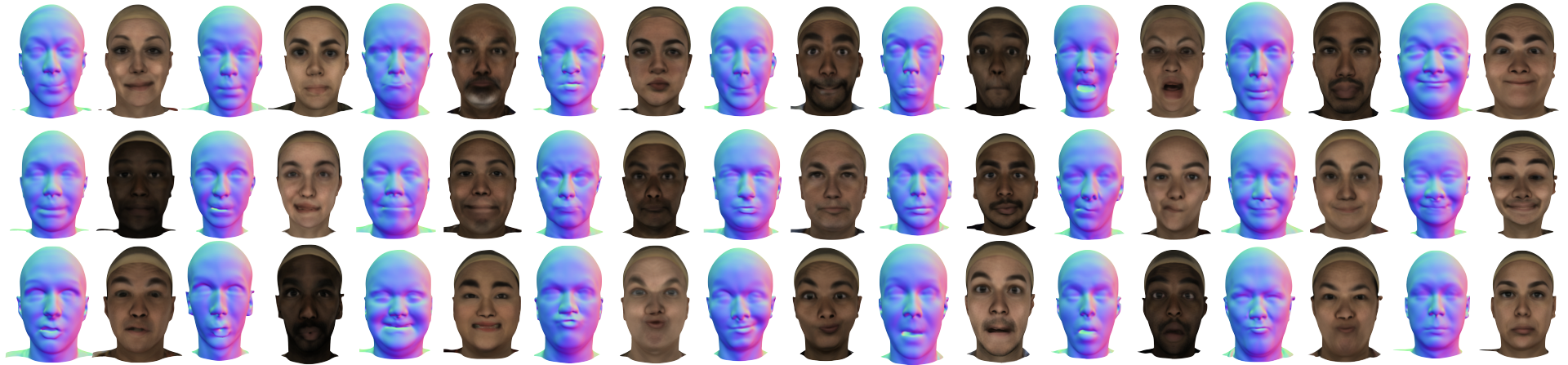}
    \caption{Random heads sampled from our model trained on the LHS dataset showing varying head shape, expression, geometric identity, and appearance. For each sample, we show 3D geometry and appearance side-by-side. }
    \label{fig:sample_buh}

\end{figure*}

The latent codes for geometric identity $\id$ and appearance identity $\appid$ are unknown before the model is trained.
We use auto-decoding \cite{i_DeepSDF} to obtain optimal latent representations.
In this set-up, training pairs of shapes and random latent codes are formed.
During training, the latent codes are updated together with the model weights.
In contrast to prior work, we additionally enforce a Gaussian prior on $\id$ and $\appid$.
This guarantees \ia that we can effectively draw samples from both latent codes during test time.
In such a setup, referred to as Variational Autodecoder (VAD) \cite{zadeh2019variational,hao2020dualsdf}, one uses the reparameterization trick from Variational Autoencoders (VAEs) \cite{kingma2013auto}: instead of directly learning the latent codes, one represents them with the parameters of the approximate posterior distribution, concretely a mean and a standard deviation vector $\vec{\mu}$ and $\vec{\sigma}$.
During training, random latent codes are sampled based on those.

We have now defined all components of our implicit head model $\globalfn$.
To train $\globalfn$, we minimize a number of loss functions, defined in the sequel, \wrt the model weights $\weights$ and the mean and standard deviation vectors of the latent embeddings $\id$ and $\appid$. %

\subsection{Losses}
For training we require pairs of 3D head scans together with imGHUM parameters approximating the scan; \cref{sec:data} gives details on how to obtain those.
We start by defining losses for the geometry, and continue with losses for the appearance, as well as regularizers.

\paragraph{Geometry.}
Given a set of points $\pointson$ on a ground truth mesh $\mesh$ describing the surface $\geo$ together with its ground truth surface normals $\gtnormal$, and corresponding imGHUM parameters $(\pose,\shape,\expression)$ approximating $\mesh$, we minimize
\begin{equation}
    \loss_{g} = \frac{1}{|\pointson|} \sum_{i \in \set{O}} \lossweight_{g}  |\dist_{\point_i}| + \lossweight_{n} \| \normal_{\point_i} - \gtnormal_i \|,
\end{equation}
where $\normal_{\point} = \nabla_{\point} \dist_{\point}$ is the estimated surface normal defined by the gradient of the estimated distance \wrt $\point$ and $\lossweight_{*}$ weight the different loss components.
Further, we supervise the SDF with the Eikonal loss \cite{gropp2020igr} for points $\pointsoff$ around the surface and additionally supervise their sign
\begin{equation}
    \loss_{e} = \frac{1}{|\pointsoff|} \sum_{i \in \pointsoff} (\| \normal_{\point_i} \| - 1)^2
\end{equation}
\begin{equation}
    \loss_{l} = \frac{1}{|\pointsoff|} \sum_{i \in \pointsoff} \text{BCE}\big(\signlabel_i, \sigmoid(\sharpness \dist_{\point_i})\big),
\end{equation}
based on binary sign labels $\signlabel$, sigmoid function $\sigmoid$, binary cross-entropy $\text{BCE}$, and the learnable parameter $\sharpness$ controlling the sharpness of the decision boundary.

\paragraph{Appearance.}
For all points, we supervise the color component of $\globalfn$ to match the ground truth color $\gtcol$ of the nearest neighbor on the surface of $\mesh$
\begin{equation}
    \loss_{c} = \frac{1}{|\pointson \cup \pointsoff|} \sum_{i \in \pointson \cup \pointsoff} | \col_{\point_i} - \gtcol_i |.
\end{equation}
While $\loss_{c}$ is theoretically sufficient to supervise the color field, in practice training only using $\loss_{c}$ leads to overly smooth results.
To this end, we propose a novel perceptual patch loss with importance sampling to improve the visual fidelity of results.
We sample a patch by first sampling a point on the mesh.
Next, we create a virtual camera viewing the sampled point along its surface normal.
Using this camera, we render a small image and additionally compute the 3D surface locations corresponding to each pixel -- see \cref{sec:data} for details.
For all 3D points in the resulting patch $\patch$ and the corresponding rendered image $\bar\image^\patch$, we enforce the previously introduced color and distance losses, and additionally apply a VGG-loss \cite{chen2017photographic} for an image $\image^\patch$ formed using the colors returned by $\globalfn$ for all $\point \in \patch$
\begin{equation}
    \small
    \loss_{p} = \frac{1}{|\patch|} \sum_{i \in \patch} \left( \lossweight_{pc} | \col_{\point_i} - \bar\image^\patch_i | + \lossweight_{pd}  |\dist_{\point_i}| \right)  + \lossweight_{pp} \text{VGG}(\image^\patch, \bar\image^\patch),
\end{equation}
where $\bar\image^\patch_i$ is the $i$-th pixel color in the ground truth rendered patch. 
In contrast to rendering losses proposed in prior work \cite{yariv2020multiview,alldieck2022phorhum}, our perceptual patch loss does not require memory or computationally expensive differentiable rendering strategies, and all the expensive operations can be performed during data pre-processing.

\paragraph{Regularization.} To enforce a Gaussian prior on $\id$ and $\appid$, we minimize the Kullback-Leibler (KL) divergence loss~\cite{kingma2013auto} for each mean and standard deviation pair $(\vec{\sigma}, \vec{\mu})$ corresponding to each instance of $\id$ and $\appid$
\begin{equation}
\loss_{\text{KL}} = -\frac{1}{2} \sum_{j=1}^{J}\left(1 + \log(\sigma_j^2) - \mu_j^2 - \sigma_j^2 \right),
\end{equation}
where $j$ references the $j$-th dimension. 

We use pairs of meshes $\mesh$ and imGHUM parameters $(\pose,\shape,\expression)$ approximating $\mesh$ during training -- see \cref{sec:data}.
Since we do not have access to ground truth imGHUM parameters, we allow our model to ``correct''  those estimates $(\pose,\shape,\expression)$ during training.
To this end, we treat $(\pose,\shape,\expression)$ as pre-initialized variables and update those together with model weights and latent representations.
We further guide the optimization using 3D facial keypoints $\landmarks$ leveraged to initialize the imGHUM parameters (\cf \cref{sec:data})
\begin{equation}
\loss_{k} = \frac{1}{|\landmarks|} \sum_{i \in \landmarks} |\dist_{\landmark_i}| + | \semantics_{\landmark_i} - \bar\semantics_i |,
\end{equation}
where $\semantics$ are reference points returned by imGHUM and $\bar\semantics$ are ground-truth reference points for the given keypoint set.

\subsection{Training Data Pre-Processing}
\label{sec:data}

Given a mesh $\mesh$ with a corresponding color texture, we fit GHUM parameters using a standard 3D facial landmark based optimization for global alignment with refinement based on Iterative
Closest Points (ICP) \cite{besl1992method}. As some of the training scans belong to the same person, we further tie the parameters of the same identity $\shape$. We collect the sets of on-surface points $\pointson$ and near-surface points $\pointsoff$ using the procedure in \cite{alldieck2022phorhum}. To gather a set of patches $\patch$, we take the following approach. First, we uniformly sample a set of points on the surface of $\mesh$. Next, we compute their distance to the closest facial landmark $\landmarks$. We transform these distances into probabilities using a radial basis function (RBF) kernel followed by normalization and draw another set of candidate samples from this distribution. By increasing the bandwidth parameter of the RBF kernel one can skew the distribution to sample more points that belong to the face region. For each point in this candidate set, we create a virtual camera looking at the sampled point along its surface normal at a fixed distance of $4$~cm.
We render the patch at a small resolution of $64\times64$~px using orthographic projection and store the color and 3D surface location of each pixel in the patch. Finally, we select from the candidate set the top patches with most coverage, \ie as few pixels associated with empty space.

\subsection{Implementation Details}
\label{sec:imp_details}
The two network branches are modeled with four 512-dimensional fully-connected layers with Swish activation \cite{ramachandran2017searching}.
We apply positional encoding \cite{tancik2020fourfeat} on $\semantics$ and $\geofeat$ before passing them to the networks.
The geometric identity $\id$ and appearance $\appid$ are learned using 128-dimensional codes, while the
$\geofeat$ is 8-dimensional.
To prevent overfitting and to obtain disentangled parameters (\cf \cref{sec:ablations}), we randomly replace $(\pose, \expression)$ pairs with Gaussian noise at $50\%$ probability.
Further, we start training with the $\loss_{\text{KL}}$ loss disabled  and then linearly increase its influence. We do not follow the auto-decoder learning scheme proposed by DeepSDF where every scan is represented in each training step. Instead, we employ the standard way of training neural networks with randomly sampled batches for each update step. See our Suppl.\ Mat.\ for training details.

\section{Experiments}
\vspace{-1mm}
\label{sec:experiments}
\begin{figure}
    \centering
    \begin{overpic}[width=0.9\linewidth]{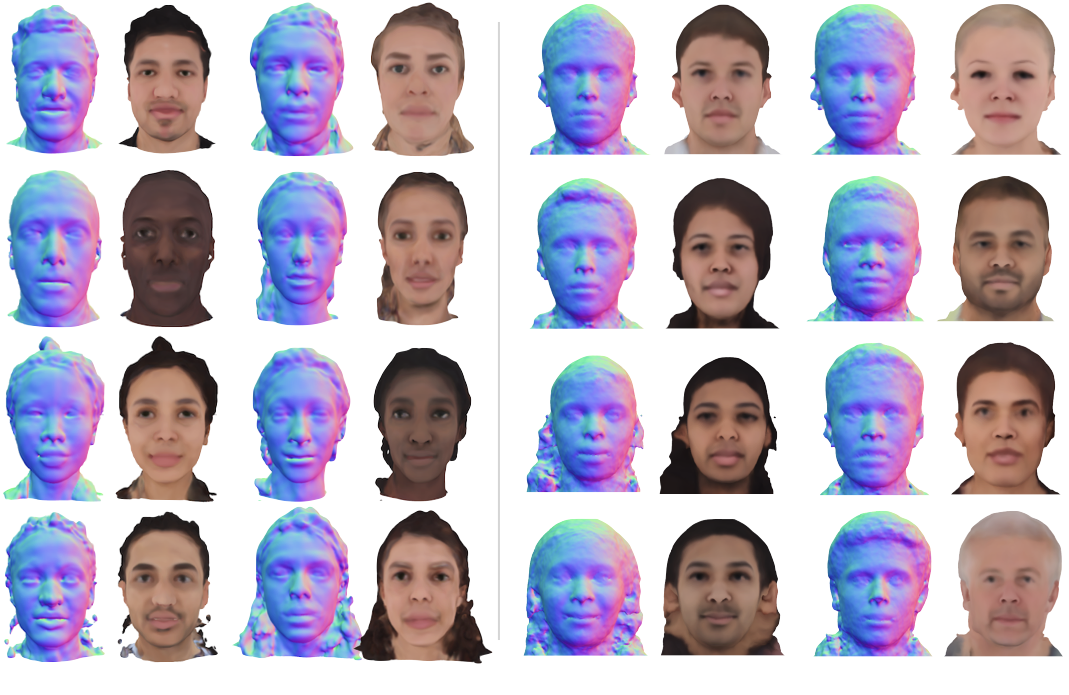} %
        \put(18,0){\footnotesize Ours}
        \put(70,0){\footnotesize i3DMM}
    \end{overpic}
    \caption{Random heads sampled from our model and i3DMM, trained on the RP dataset. In contrast to i3DMM, our model always matches geometry and appearance. Additionally, our results are more detailed both in geometry and color. Note that we visualize the raw, unshaded results.}
    \label{fig:sample_rp}

\vspace{5mm}
    \centering
    \includegraphics[width=\linewidth]{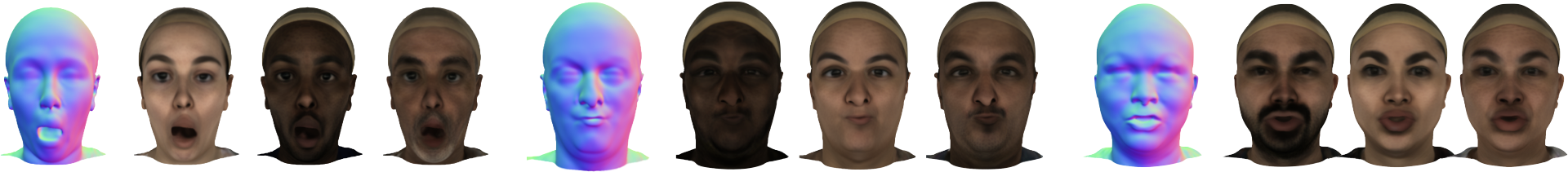}
    \caption{All factors in our model are disentangled allowing for fine-grained control. Here we sample three different matching appearances for each subject, while keeping remaining factors (geometric identity, head shape, pose, expression) fixed.}
    \label{fig:vary_appearance}
\end{figure}

We evaluate our model and its design choices by means of ablations and comparisons. In \cref{sec:rep_power} we demonstrate the representation power and the properties of our model.
We continue by comparing PhoMoH with i3DMM, a recent implicit 3D model of human heads, and ablate critical design choices.
Finally we illustrate several applications enabled by our model.

\paragraph{Datasets.}
We train and evaluate our model on two different datasets with different characteristics.
First, we rely on a dataset captured in the lab, under controlled settings. We refer to this dataset as LHS (LabHeadScans).
LHS includes scans of $200$ individuals taking up to $40$ different combinations of facial expressions and head poses each, totaling in $7,839$ scans. All subjects have given informed consent covering the use-case presented here.
Since this dataset was originally captured for building a classical mesh-based head model, the subjects are wearing caps and variation in hair styles is lacking. However, facial hair is present in the data.
To demonstrate the full potential of our model, we additionally train it on as little as $100$ scans purchased from RenderPeople~\cite{renderpeople}, which we refer to as RP.
By evaluating our method on both datasets, the rich LHD dataset and the very small but diverse RP dataset, we are able to showcase different properties and the unique strengths of PhoMoH.

\paragraph{Baselines.}
We compare our model with i3DMM~\cite{yenamandra2021i3dmm}, a recent implicit 3D model of human heads.
In contrast to our model, i3DMM is not extending an existing head model but is learned directly from scans.
i3DMM allows to control for identity, appearance, and facial expression -- we additionally allow control over head pose and shape. 
i3DMM's latent codes are unstructured, and thus do not allow for direct sampling.
Further, i3DMM expression code is not coupled with any other model and is usually controlled by moving along the directions of the training expressions.
Finally, i3DMM learns a reference shape to disentangle color from appearance. However, this leads to problems in practice, as we will show.
i3DMM requires to compute gradients \wrt to all scans in each training step, which makes it slow and memory-intensive to train.
To be able to train i3DMM we therefore had to define a new dataset LHS-40, a subset of LHS including only 40 of 200 individuals.
We use LHS-40 or RP whenever comparing with i3DMM.
We further compare against imGHUM~\cite{alldieck2020imghum} and FLAME~\cite{FLAME:SiggraphAsia2017} for the task of mesh reconstruction, and to show how PhoMoH adds benefits in representation power.

\subsection{Representation Power}
\label{sec:rep_power}
\begin{figure}

    \centering
    \includegraphics[width=\linewidth]{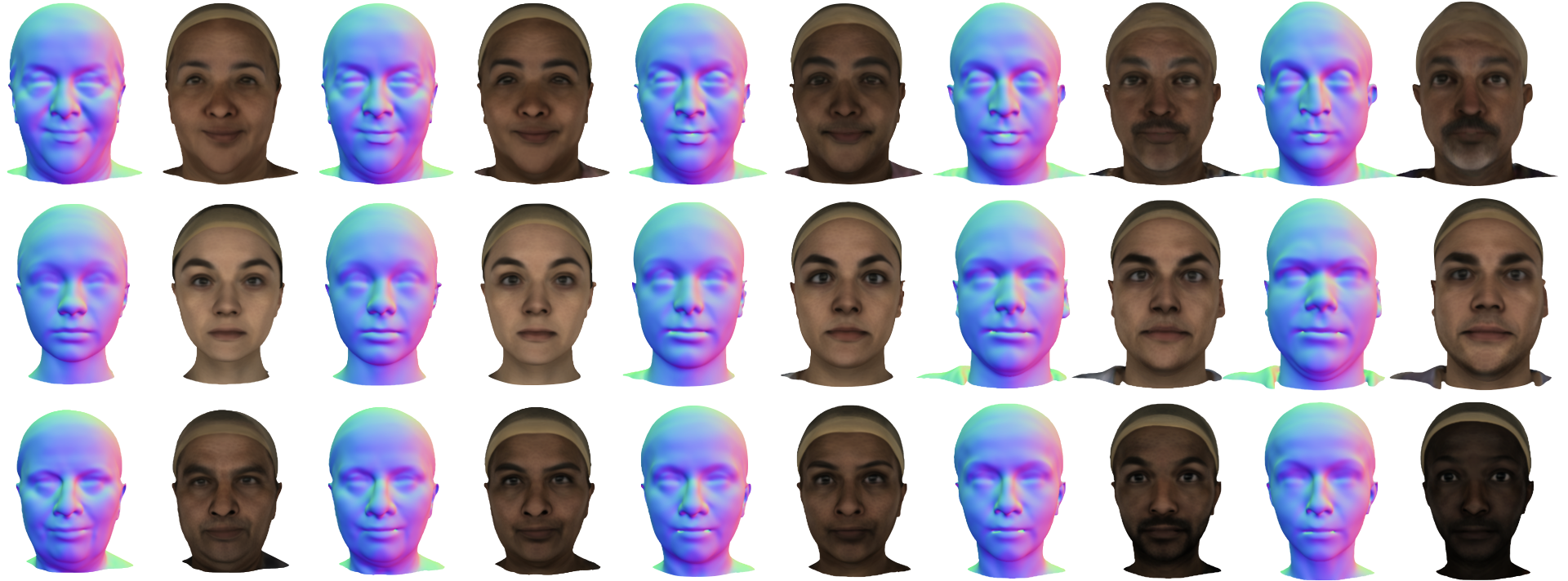}
    \caption{By enforcing a Gaussian prior on latent codes, we ensure compact, well covered latent spaces. We demonstrate this property by linearly interpolating three configuration pairs (left, right) to create plausible intermediate identities (middle).}
    \label{fig:interpolate}
\end{figure}

We demonstrate the representation power of PhoMoH by randomly sampling from the latent embeddings for identity, appearance, head shape, and expression. 
In \cref{fig:sample_buh} and \cref{fig:sample_rp} we show results for both datasets.
See the Suppl.\ Mat.\ for additional results and a numerical comparison.
The ability to directly sample from the latent embeddings is enabled by the Gaussian prior imposed during training.
In contrast, for i3DMM one has to run PCA on the latent spaces to support sampling.
Thus, sampling is performed without knowledge of the true underlying data distribution.
In contrast to our results, outputs by i3DMM are less detailed and geometry and color are not always consistent.
Additionally, i3DMM is learned in a normalized space, thus all samples have ambiguous scale.
In contrast, PhoMoH is defined in a metric space.
Geometric identity and appearance are decoupled in our model.
This allows us to sample multiple plausible appearances for a given geometry, as shown in \cref{fig:vary_appearance}.
Finally, we interpolate between two sampled model configurations in \cref{fig:interpolate}.
The intermediate results are detailed and plausible, demonstrating a compact, well-covered latent space.

\subsection{Mesh Reconstruction}
\label{sec:mesh_recon}
\begin{figure}
    \centering
    \includegraphics[width=\linewidth]{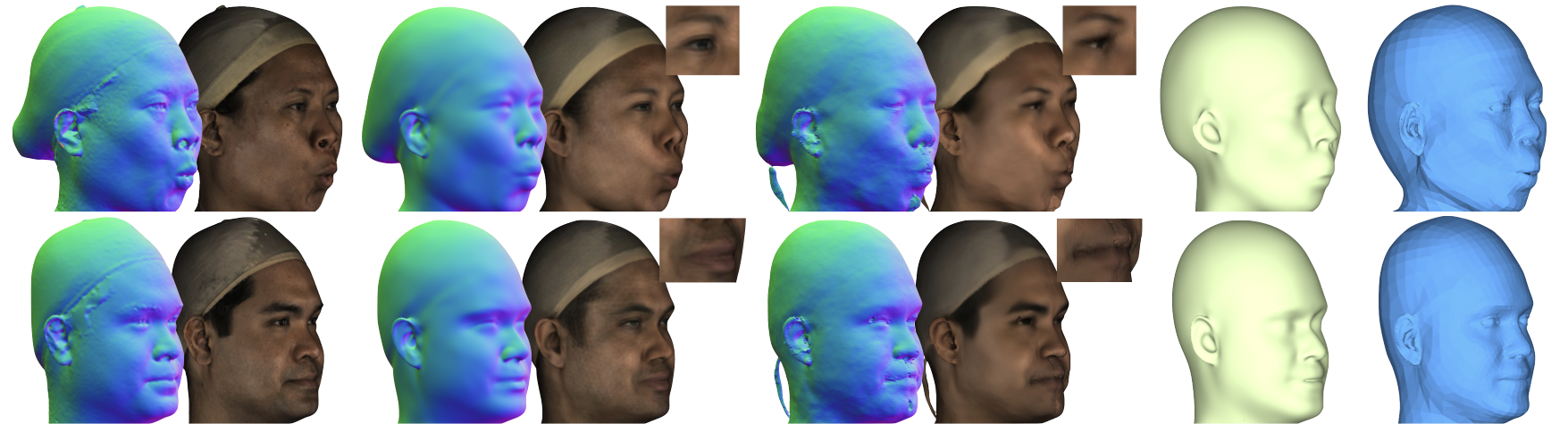}
    \caption{Qualitative comparison of fitting results. From left to right: target scan, ours, i3DMM, imGHUM, and FLAME.}
    \label{fig:fitting}
\end{figure}

One possible use case of PhoMoH, among others, is representing a scan by fitting our model to it.
We compare our model on this task with i3DMM, FLAME, and imGHUM using the test set of our LHS dataset.
For fair comparisons with i3DMM, as mentioned previously, we created a separate train split, LHS-40, which includes only 40 of the original scan identities.
We trained both our model and i3DMM on this smaller subset. 
The test split contains 10 unseen identities with up to 40 expressions each and 366 scans in total. 
We pre-process the test split for i3DMM (canonicalizing pose, normalizing scale) and use the raw scans for all other methods.
We report bidirectional Point to Surface Distance (P2S) in mm, Chamfer distance ($\times 10^{-3}$), and Normal Consistency (NC) for geometry reconstruction, and P2S color difference for assessing the color reconstruction quality. Additionally, we report the Learned Perceptual Image Patch Similarity \cite{zhang2018unreasonable} (LPIPS) on frontal renderings, see \cref{tbl:fitting}. Our geometric reconstructions are numerically more accurate than i3DMM as this occasionally features floating artifacts and discontinuities caused by their reference shape strategy. i3DMM has numerically slightly better P2S color quality, but the results appear blurry, as evident in \cref{fig:fitting} and from LPIPS results. We include imGHUM in this comparison demonstrating that most of the detail in the geometric reconstruction is recovered by the geometric identity layer of our method. imGHUM is a good proxy, but is restricted to a coarse representation of a scan. Further, we compare against FLAME, a popular mesh-based 3DMM. FLAME is high-dimensional with a shape space of 300 PCA bases and an expression space of 100 bases. In contrast, imGHUM, the model PhoMoH shares its base parameterization with, is controlled by 16 and 20-dimensional shape and expression codes, respectively. Nevertheless, FLAME produces less personalized results than PhoMoH which is evident in lower NC and P2S. FLAME reports the best results on Chamfer. This is likely because PhoMoH and imGHUM always produce a physically plausible oral cavity, even if not present in the fitted scan. This affects the Chamfer distance the most as this measures the quadratic point to point error. In summary, our method produces the best overall results and is the only method capturing sharp and realistic appearance.

\begin{table}
\centering
\small
\resizebox{0.99\linewidth}{!}{%
\begin{tabular}[t]{l||c|c|c||c|c}
 & P2S $\downarrow$ & Ch. $\downarrow$ & NC $\uparrow$ & Color  $\downarrow$ & LPIPS $\downarrow$ \\
\hline 
i3DMM \cite{yenamandra2021i3dmm} & 2.009 & 0.322 & \textbf{0.966} & \textbf{0.057} & 0.140 \\
imGHUM \cite{alldieck2020imghum} $\dagger$ & 2.957 & 0.121 & 0.946 & -- & --\\
FLAME \cite{FLAME:SiggraphAsia2017} & 1.767 & \textbf{0.029} & 0.945 & --& --\\
\hline
Ours (LHS-40) $\dagger$ & \textbf{1.425} & 0.034 & \textbf{0.966} & 0.069 & \textbf{0.125}\\

\end{tabular}
}
\caption{Fitting results on the LHS-40 test set. Methods with $\dagger$ produce an oral cavity not present in the scans, which negatively impacts the numerical results, see discussion.}
\label{tbl:fitting}
\end{table}

\subsection{Ablations}
\label{sec:ablations}
We ablate several important design choices: the introduction of the patch loss $\loss_p$, strategies for $(\pose, \expression)$ conditioning, and the normalization of the geometric feature $\geofeat$.

\begin{figure}
    \centering
    \begin{overpic}[width=1.0\linewidth]{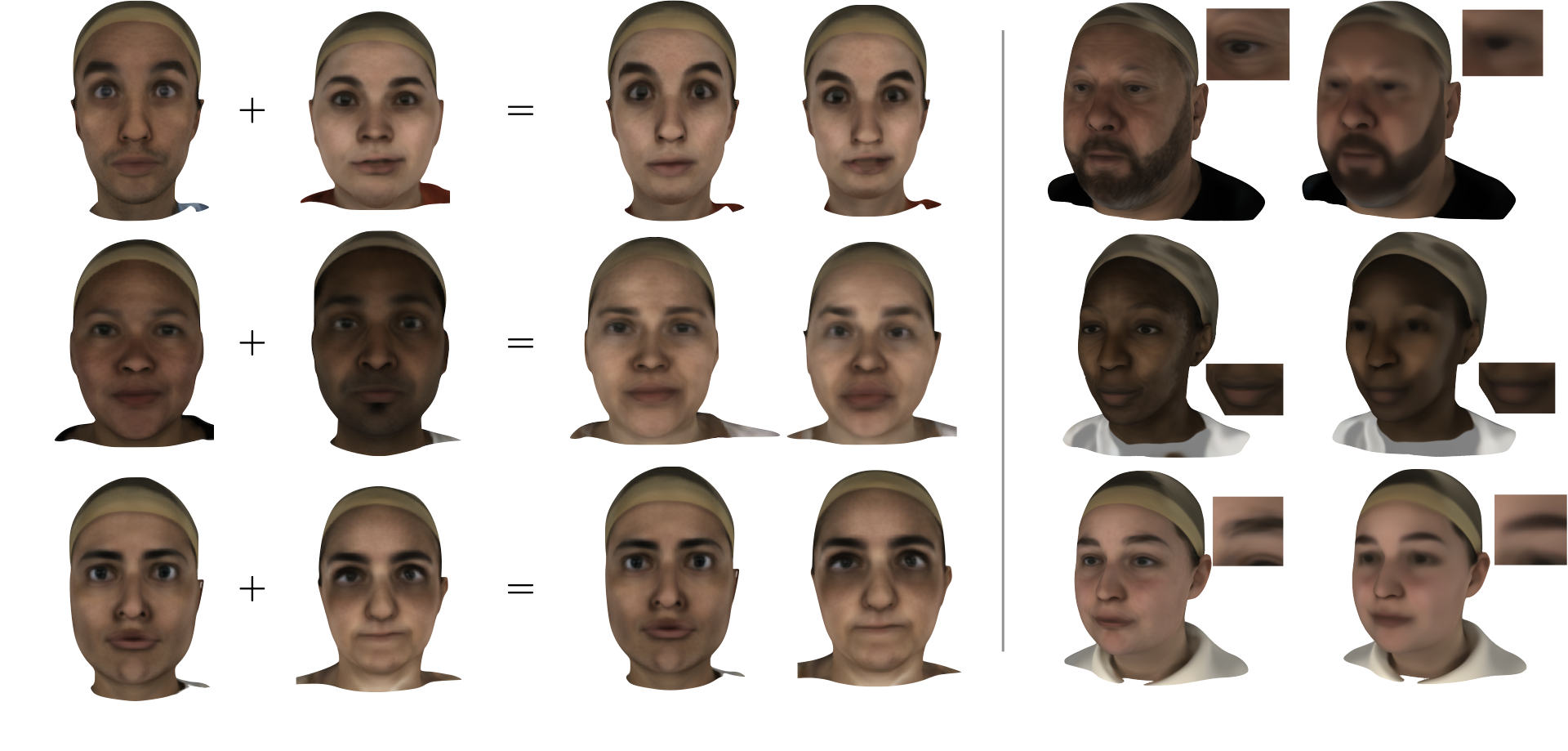} %
        \put(67,0){\scriptsize Full Model}
        \put(88,0){\scriptsize $-\loss_p$}
        \put(4.5,0){\scriptsize Source}
        \put(20.5,0){\scriptsize Target}

        \put(41,0){\scriptsize (A)}
        \put(54,0){\scriptsize (B)}

        \put(0,34){\scriptsize \rotatebox{90}{Full Model}}
        \put(0,19){\scriptsize \rotatebox{90}{$-LN(\geofeat)$}}
        \put(0,1.5){\scriptsize \rotatebox{90}{$-\text{rand}(\pose, \expression)$}}
    \end{overpic}
    \caption{Ablation of critical design choices. Left: Layer normalization and randomization are essential to obtain disentanglement of model parameters. We show the source head with target appearance (A) and 
    target appearance/pose/expression (B). Only our full model produces the desired result. Right: Our novel patch loss $\loss_p$ significantly increases the visual fidelity of results.}
    \label{fig:ablate}
\end{figure}
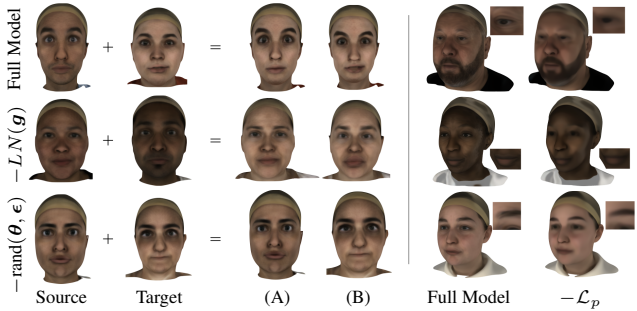

Our proposed patch loss $\loss_p$ significantly improves the visual fidelity of the results. See \cref{fig:ablate} (right) for a side-by-side comparison. This is also evident from a significantly lower Frechet Inception Distance \cite{salimans2016improved} (FID) compared to a model trained without $\loss_p$. To compute the FID score ($\downarrow$) we rendered samples of each model and compare them against renderings of real scans. Our full model has a FID score of 50.80 whereas a model without $\loss_p$ has a score of 80.69. %

Randomly replacing pose and expression $(\pose, \expression)$ pairs with Gaussian noise during training is crucial for disentangled latent representations. When $(\pose, \expression)$ are always input to $\geofn$ and $\colfn$, the networks can choose to ignore the geometry and appearance inputs $\id$ and $\appid$ and partly or entirely and correlate shape and appearance with $(\pose, \expression)$ instead. Not using $(\pose, \expression)$, on the other hand, degrades visual fidelity (FID: 61.19). Failed disentanglement can also be observed when not applying layer normalization on the geometric feature $\geofeat$. We hypothesize that, without constraining the distribution of $\geofeat$, the network is able to leak geometric information into the appearance network which prevents proper disentanglement of $\id$ and $\appid$. In \cref{fig:ablate} (left), we show examples of such failed disentanglement along with results from our model. The appearance code lost its meaning for the example without layer normalization (line 2). Instead of applying the target appearance, the model produces a random (yet realistic) result. For the experiment without $(\pose, \expression)$ randomization (line 3), we observe that appearance is correlated with pose and expression instead of being controlled with $\appid$.
In contrast, our compete model produces the desired result.

\begin{figure}
    \centering
    \begin{overpic}[width=0.95\linewidth]{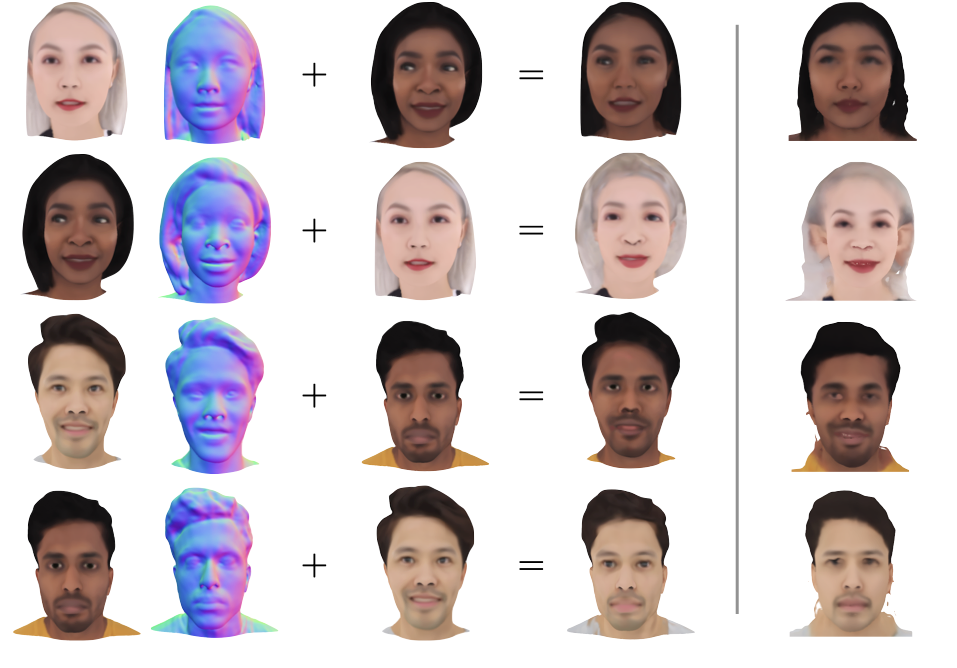} %
        \put(11,0){\footnotesize Source}
        \put(40,0){\footnotesize Target}
        \put(63,0){\footnotesize Ours}
        \put(83.5,0){\footnotesize i3DMM}
    \end{overpic}
    \caption{The proposed geometric feature enables semantically coherent appearance transfer between samples, useful for \eg creative editing or dataset augmentation and diversification. From left to right: original appearance together with its geometric identity, target appearance, the result of appearance transfer, result of i3DMM~\cite{yenamandra2021i3dmm}. Please note semantically coherent results of our method for \eg hair coloring, despite different hair styles between source and target. i3DMM fails to preserve the hair line, as it is especially visible in the 2nd row (ear region).}
    \label{fig:app_transfer}
\end{figure}

\begin{figure}
    \centering
    \includegraphics[width=0.95\linewidth]{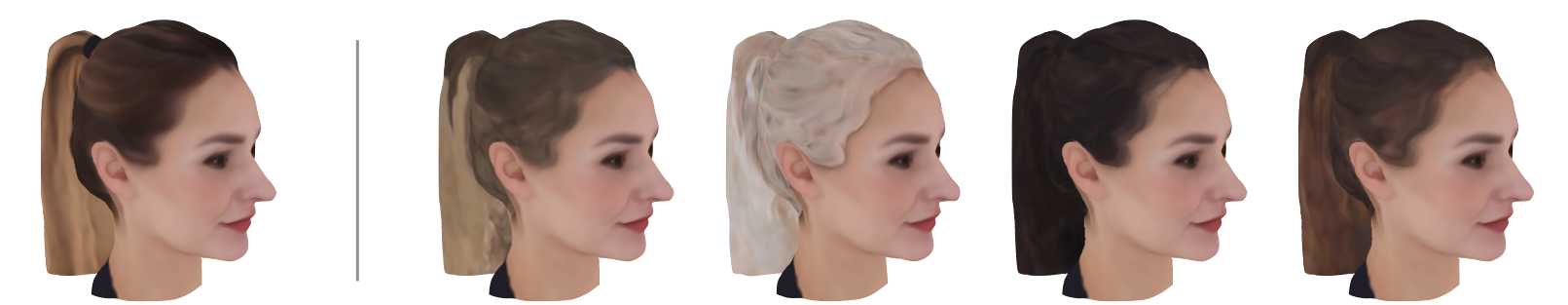}
    \caption{By freezing the appearance code for some portions of the reconstructed mesh, and re-querying other parts, we can perform semantically coherent appearance edits. Here we sample new colors for points corresponding to the hair in the source mesh (left). Note how the model produces plausible novel hair colors.}
    \label{fig:edit_hair}
\end{figure}

\begin{figure}
    \centering
    \includegraphics[width=0.95\linewidth]{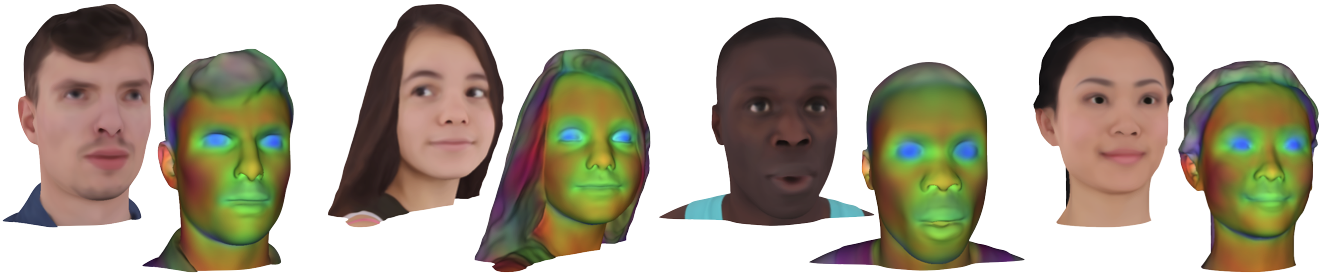}
    \caption{Our method learns a geometric feature $\geofeat$ (without explicit supervision) to serve as semantically informed input to the appearance network. $\geofeat$ is a useful tool beyond its original use case as it defines dense correspondences between all sampled heads. Here we show the first 3 PCA components as RGB colors.}
    \label{fig:geo_feature}
\end{figure}

\begin{figure}
    \centering
    \includegraphics[width=0.95\linewidth]{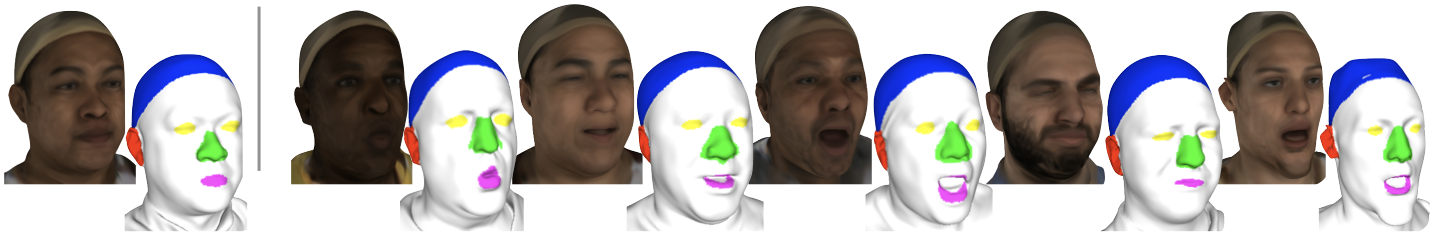}
    \caption{We transfer semantic labels from one head (left) to others (right) by training a classifier based on the geometric feature $\geofeat$. The classifier correctly predicts semantic labels for diverse heads despite being trained on a single head instance.}
    \label{fig:one_shot}
    \vspace{-1mm}
\end{figure}

\subsection{Applications}
\label{sec:applications}
Finally, we are demonstrating additional use-cases of PhoMoH.
The per-point geometric feature $\geofeat$ learned by our model can be understood as semantic correspondences between different head instances, see \cref{fig:geo_feature}.
This is a useful property, as we now show.
First, $\geofeat$ enables semantically correct appearance transfer between different head instances, useful for \eg dataset augmentation or creative editing.
We show results in \cref{fig:app_transfer} and compare to appearance transfer results obtained using i3DMM.
While our model correctly places colors semantically, i3DMM struggles to maintain the hair line defined by the geometry.
This is especially apparent in row two, where i3DMM textures the hair region with colors from the ear region.
Similar problems can be observed in \cref{fig:sample_rp}.
i3DMM's inability to follow the semantics defined by the geometry is rooted in model's design, where learning of color is decoupled from geometry on a canonical reference shape.
Our model also supports local appearance editing, as shown in \cref{fig:edit_hair}.
In this experiment, we re-queried points corresponding to hair with newly sampled appearance codes.
Our model produces plausible hair colors, following the shape defined by the original hairstyle.
The geometric feature is also useful for classification and segmentation tasks.
In our final experiment, we labeled one single head instance with six semantic labels, corresponding to cap, eyes, ears, nose, mouth, and skin.
We then trained a classifier mapping $\geofeat$ to one of the six labels.
Despite being trained only on a single head, our classifier gracefully generalizes to samples produced by PhoMoH, see \cref{fig:one_shot}.
This means, once trained, PhoMoH can drastically simplify and speed up the labeling process for a given dataset.

\section{Discussion \& Conclusions}
\vspace{-1mm}
\label{sec:conclusion}
\paragraph{Limitations.}While already modeling large parts of the human head, our model still lacks certain features for true realism. For instance, our model does not capture dynamics that may be present \eg in the hair caused by motion or gravity. Furthermore, our model currently does not allow to control eye-movements and extreme expressions are somewhat limited by the expressiveness of imGHUM. Finally, PhoMoH occasionally produces blurry appearance or geometry defects.

\paragraph{Ethical Considerations.}Our model is a useful tool for generating diverse instances of human heads, even when trained on relatively little data.
Its unique properties allow for disentangled control over
head geometry and semantically coherent appearance.
Our model is not intended or particularly useful for any form of deep fakes.
In order to \emph{faithfully} represent an existing person with our model, we would need access to high-quality 3D scans of that person.
While our model can be animated and realistically rendered (without audio), there is still a realism gap between our renders and real video (see limitations).

\paragraph{Conclusions.}We have presented PhoMoH, a novel methodology to construct photorealistic generative 3D head models with disentangled control over geometric identity, appearance, expression, head pose, and head shape.
In contrast to prior work, our implicit formulation makes it possible to model the human head as a whole, including hair and clothing.
An important property of PhoMoH are the proposed per-point geometric features learned without explicit supervision.
The geometric features not only ensure that appearance is coherent with the semantics defined by the geometry, but they additionally enable interesting applications such as semantic label transfer or appearance editing.
In the future, we would like to add control over eye-movements as well as allow for decoupled control over the semantic regions of the head, like, for instance the hair style. 

\appendix

\section*{Supplementary Material}

In this supplementary material, we detail our implementation, list the values of all hyper-parameters, give inferences timings, and present additional results and experiments. Please also see the accompanying video.

\section{Implementation Details}

The two network branches, for geometry and color, share the same architecture represented by $4$ Dense layers with Swish activation \cite{ramachandran2017searching} as the non-linearity. The hidden size of a Dense layer is set to $512$. We set the size of both our geometric identity code $\id$ and appearance $\appid$ to $128$, and the size of the geometric feature code $\geofeat$ to $8$. We apply 5 and 10-dimensional positional encoding \cite{tancik2020fourfeat} on $\semantics$ and $\geofeat$, respectively, before passing them to the networks. In total, our architecture has $\approx1.8$M trainable parameters.

We train for $1.5$M steps, using $2$ Adam optimizers, one for the network weights and one for the identity codes, respectively. For the former, the learning rate is set to $5\times10^{-4}$, while for the latter to $1\times10^{-3}$, with an exponential decay schedule (i.e. $0.8$ decay every $5000$ steps).
For RP and LHS-40, we train only for $150$k steps, to prevent overfitting on those small datasets.
We train with a batch size of $256$ on 16 TPUv2. Each training example consists of $512$ points sampled on surface and $2\times512$ around the surface. For the latter, we consider $512$ samples of points closer to the surface and additional $512$ points uniformly sampled in the scene bounding box. We also include one randomly selected $64\times64$~px patch out of $1024$ pre-computed patches for each training example in the batch. Our training weights for the losses are set to the following values: $\lambda_n=1.0$, $\lambda_e=0.1$, $\lambda_l=0.1$, $\lambda_c=2$, $\lambda_{pc}=1.0$, $\lambda_{pd}=5.0$, $\lambda_{pp}=0.3$, $\lambda_{k}=10.0$. Further, we start training with the $\loss_{\text{KL}}$ loss disabled and then linearly increase it to $\lambda_{kl}=0.05$ over the course of the first $200$k training steps. Similarly, we start with $\lambda_g=5.0$ and linearly increase it to $\lambda_{g}=20.0$ over $50$k steps.

To prevent our networks to disregard the information in the identity codes and overfit on the GHUM parameters, we replace $(\pose, \expression)$ pairs passed to $\geofn$ and $\colfn$ with Gaussian noise with $50\%$ probability. For RP, where no per subject $(\pose, \expression)$ variation is present, we disable conditioning on those factors. Finally, we also optimize the GHUM parameters together with the rest of our architecture. %

\section{Inference Timings}

During inference, we run Marching Cubes over the distance field defined by the geometry network $\geofn$. We first discretize and probe the 3D space (represented as a rectangular cuboid centered at the origin) at a coarse resolution and then, by employing Octree sampling, we gradually raise the resolution as we come nearer to the surface. This is highly effective at reducing the computational load. We query in batches of $64^3 / 2$ points. As Marching Cubes produces a triangular mesh, our next step is to use its vertices to query our full network $\globalfn$ in order to obtain per-vertex color $\col$. The reconstruction of a colored mesh at a target resolution of $256^3$ takes on average $1.13$ seconds on an NVIDIA V100. When using a very fine resolution of $512^3$, the time needed increases to $5.25$ seconds on average. %
The resultant mesh can be imported and rendered using any conventional 3D graphics software or library (e.g. Blender, Meshlab, Unreal Engine, Unity).

\begin{table}
    \begin{center}
    \small
    \resizebox{\linewidth}{!}{%
    \begin{tabular}{cccccccc|l}
    \rot{Head model} & \rot{Metric} & \rot{Template free (implicit)} & \rot{Unstructured training data} & \rot{Pose control} & \rot{Appearance} & \rot{Hair, Clothing}  & \rot{Oral cavity} \\
    \hline\hline
    \cmark & \xmark & \cmark & \cmark & \xmark & \cmark & \cmark & \xmark & i3DMM \cite{yenamandra2021i3dmm} \\
    \cmark & \cmark & \cmark & \xmark & \cmark & \xmark & \xmark & \cmark & imGHUM \cite{alldieck2020imghum} \\
    \cmark & \cmark & \xmark & \xmark & \cmark & \xmark$^\dagger$ & \xmark & \xmark & FLAME \cite{FLAME:SiggraphAsia2017} \\
    \xmark & \xmark$^\ddagger$ & \cmark & \xmark & n/a & \xmark & \xmark & \cmark & ImFace \cite{zheng2022imface} \\
    \xmark & \cmark & \xmark & \xmark & n/a & \cmark & \xmark & \xmark & BFM \cite{gerig2018morphable} \\
    \hline
    \cmark & \cmark & \cmark & \cmark & \cmark & \cmark & \cmark & \cmark & \textbf{PhoMoH} \\
    \end{tabular}
    }
    \end{center}
    \vspace{-2.5mm}
    \footnotesize{$^\dagger$ extension for face region proposed in \cite{smith2020morphable}, $^\ddagger$ uniformly scaled space}
\vspace{1mm}
    \caption{Overview of different features of PhoMoH and all baselines. PhoMoH is the most flexible and complete model.}
    \label{tab:features}
\end{table}

\begin{figure*}
    \centering
    \vspace{2cm}
    \includegraphics[width=\linewidth]{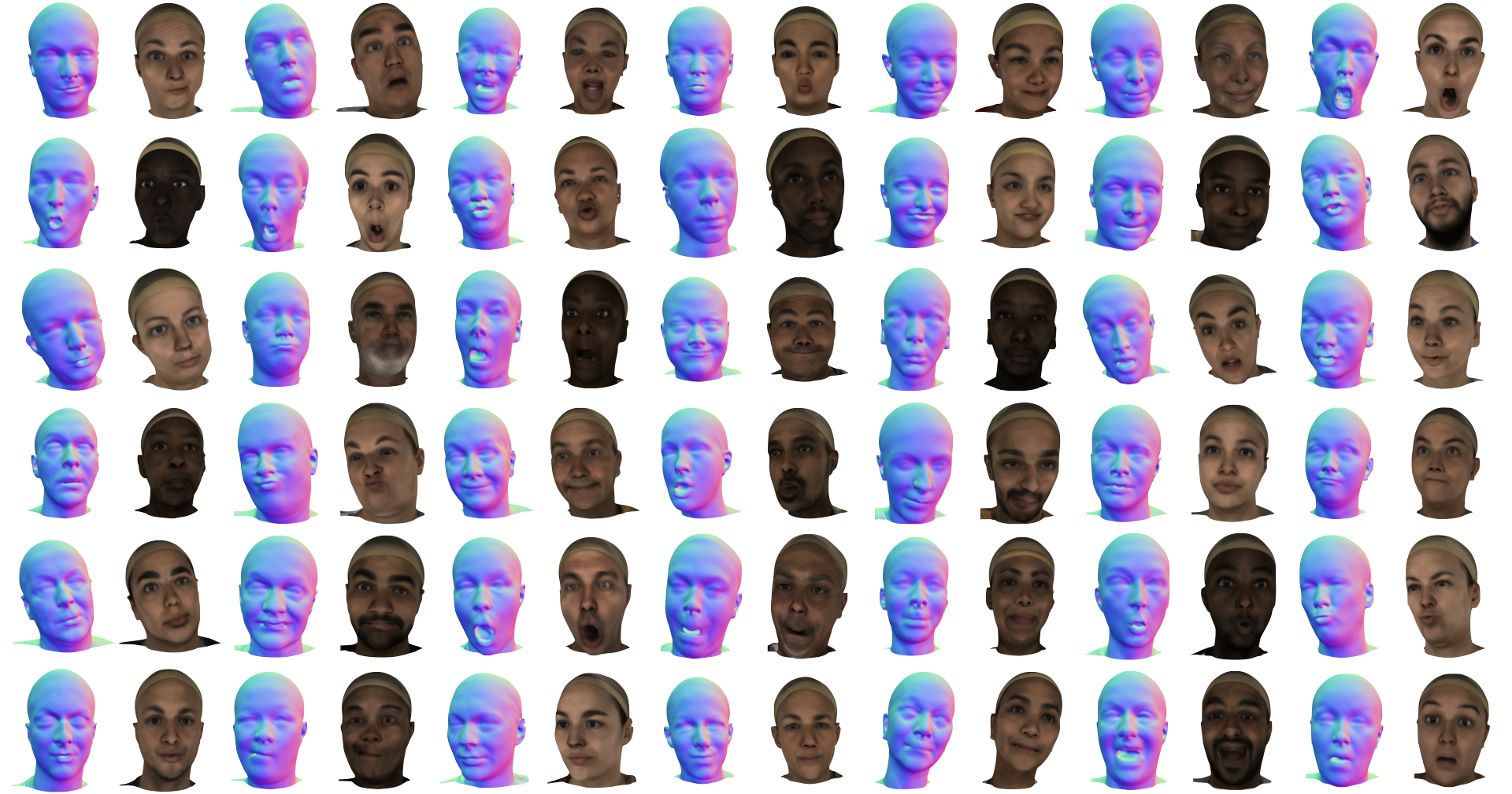}%
    \vspace{5mm}
    \includegraphics[width=\linewidth]{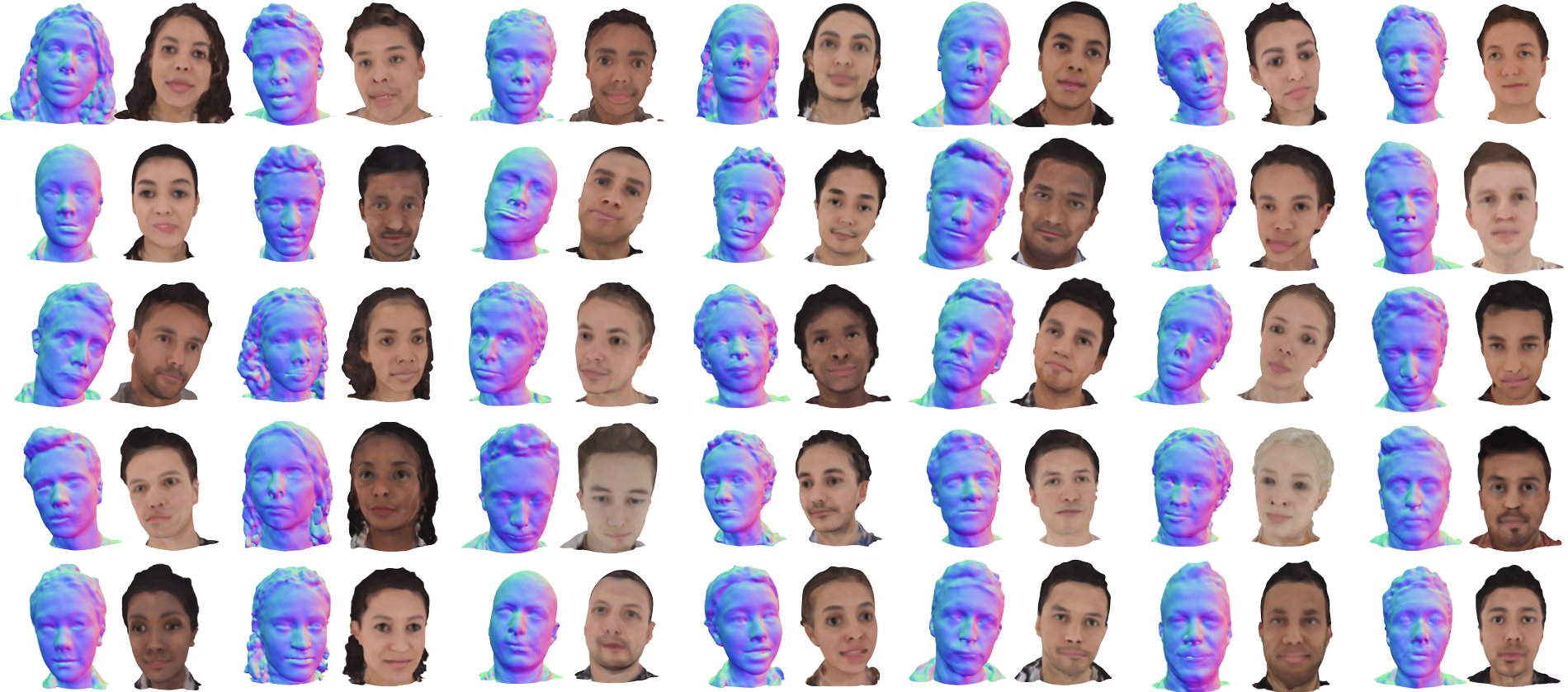}%
    \vspace{2mm}
    \caption{Random heads sampled from our models trained on the LHS (top) and RP (bottom) datasets varying all factors, namely geometric identity, appearance, expression, head pose, and head shape. For each sample, we show 3D geometry and appearance side-by-side. }
    \label{fig:sample_all}
    \vspace{2cm}
\end{figure*}

\section{Additional Results \& Experiments}

\begin{figure*}
    \centering
    \includegraphics[width=0.8\linewidth]{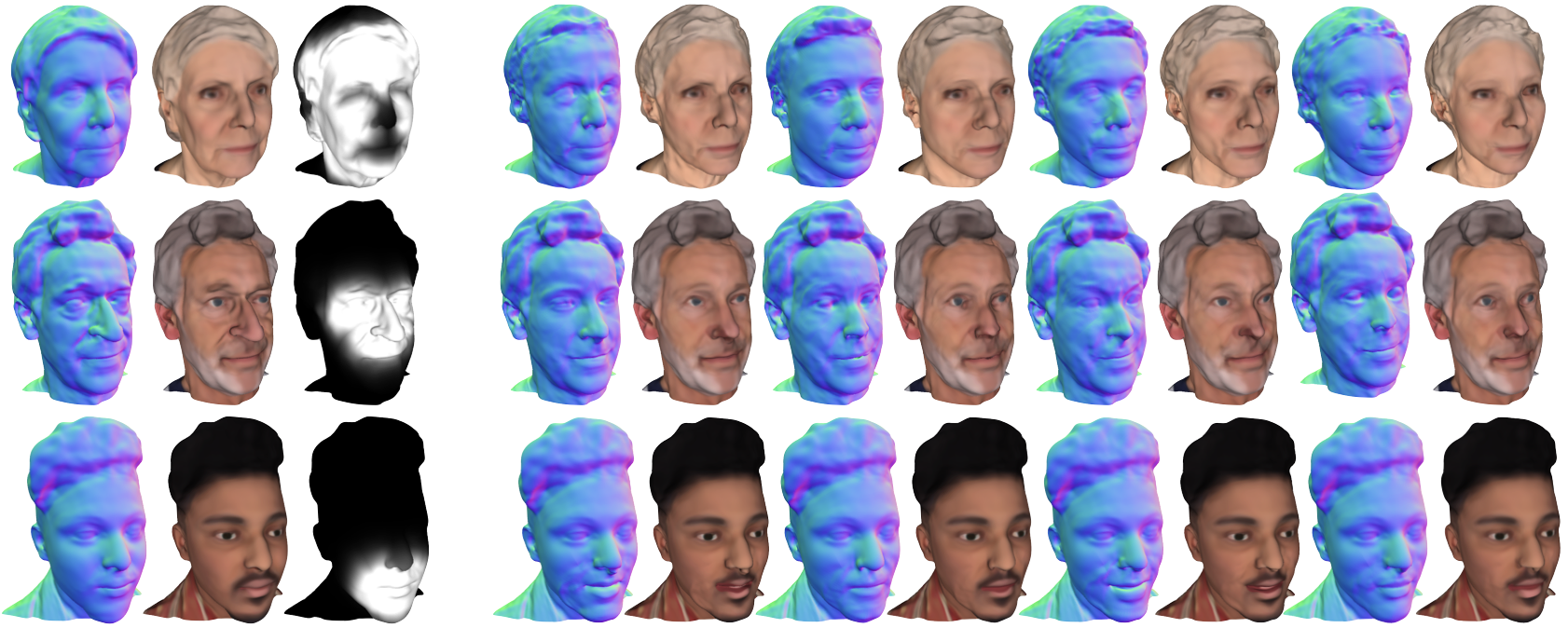}
    \caption{Local geometry editing: We define local regions based on imGHUM reference points $\semantics$ and sample new geometric identities for points belonging to those regions. From left to right: Original geometry, appearance, region labels, four geometry and appearance pairs with locally altered geometry.}
    \label{fig:local_edit}
\end{figure*}

In the following, we show additional random samples taken from our models, demonstrate the usage of our model for dataset creation, and present a strategy for local geometry edits.

\subsection{Random Head Sampling}
In \cref{fig:sample_all}, we show more randomly sampled heads using our models.
In contrast to Fig.\ 3 and Fig.\ 4 in the main paper, we additionally randomize the head pose making full use of all five control parameters of our model: geometric identity, appearance, expression, head pose, and head shape.
Please note that the RP dataset contains very little variation in head pose and expression, yet our model is able to produce heads with these factors varied.

Next, we numerical compare the quality of samples of our method and samples produced by our main baseline i3DMM.
For a fair numerical comparison the same (ideally statistical) strategy for both methods would be required. Unfortunately, this is not possible: We sample from PhoMoH using its latent statistical model. i3DMM's latent spaces do not define a statistical model,
so ``sampling'' from it requires ad-hoc solutions (trading-off between quality and diversity).
Nevertheless, we attempt a fair comparison here by hand-tuning sampling parameters per PCA-component for i3DMM's sampling strategy and transforming its results back into the metric space (results are in a canonical, non-metric space).
As i3DMM cannot be trained with our full dataset (as explained in \cref{sec:experiments} Baselines) and less data leads to degraded sampling performance, we compare here with our model trained on the smaller dataset LHS-40. Like in the main paper, we rendered samples of each model and compare them against renderings of real scans to compute the FID score ($\downarrow$). Our LHS-40 model has a FID score of $68.73$ whereas i3DMM has a score of $81.37$. While we stress again that numbers vary based on the exact sampling strategy, we conclude that our model produces more realistic images. This also correlates well with the qualitative comparisons we made in main paper. Further, our model can be trained with much more data resulting in even improved performance.

\subsection{Local Geometry Editing}

The geometric identity component of PhoMoH is modeled as a layer around imGHUM.
In the process of computing the signed distance of a point in space $\point$,
we compute its nearest neighbor $\semantics$ on the imGHUM surface.
By defining local regions based on $\semantics$, we can locally edit the geometric identity.
In \cref{fig:local_edit}, we defined three semantic regions illustrated by the white label.
We then we edited three scans embedded in our model, by randomly sampling geometric identity codes $\id$ for points $\point$ with the nearest neighbor of $\semantics$ falling into the defined region and keeping the original code otherwise.

\subsection{ImFace Comparison}

\begin{figure}
    \centering
    \includegraphics[width=\linewidth]{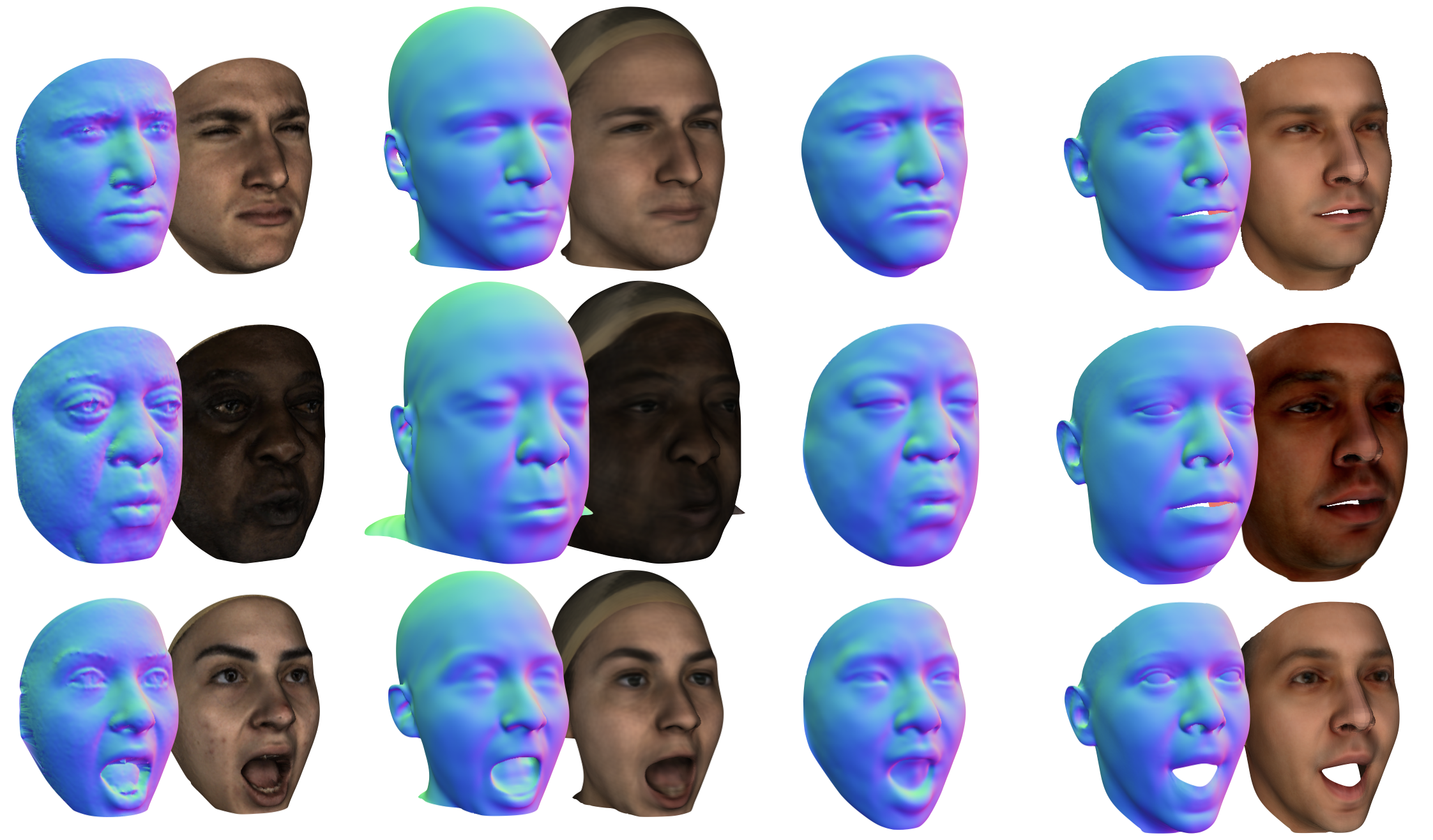}%
    \caption{Qualitative comparison of face fitting results. From left to right: ground truth face scan, ours, ImFace, and BFM. PhoMoH is on par with the specialized face model ImFace in terms of geometry reconstruction and is additionally able to reconstruct  color and to realistically complete the full head.}
    \label{fig:fitting_imface}
\end{figure}

\begin{table}
\centering
\small
\begin{tabular}[t]{l||c|c||c}
 & S2M $\downarrow$ & UNC $\uparrow$ & Color $\downarrow$ \\
\hline 
ImFace \cite{zheng2022imface} & 0.618 & \textbf{0.964} & - \\
BFM \cite{gerig2018morphable} & 1.033  & 0.948  &  0.183 \\
\hline
Ours &  \textbf{0.604} &  0.956 & \textbf{0.049}  \\

\end{tabular}
\caption{Numerical results of the face fitting experiment.}
\label{tab:fitting_face}
\end{table}

ImFace \cite{zheng2022imface} is an implicit 3D morphable face model with separate control over identity and expression. ImFace was trained on registered meshes from FaceScape \cite{yang2020facescape} with shared topology and in a canonicalized pose.
In contrast, PhoMoH can be trained with unstructured scans with varying pose.
More importantly, ImFace only models the \emph{face geometry}, whereas PhoMoH models both the geometry and appearance of the \emph{whole head} -- a significantly harder task.
A overview of different features of PhoMoH and all compared models is given in \cref{tab:features}.
Despite the different features and capabilities, we compare here with ImFace on the task of 3D fitting to scans, similar to \cref{sec:mesh_recon} in the main paper.
For this comparison, we processed our LHS test scans using ImFace's cropping and pseudo watertighting algorithm. We fit both ImFace, using the official implementation, and PhoMoH to the processed scans.
In contrast to the fitting experiment in the main paper, here we report unidirectional Scan to Mesh distance (S2M) and unidirectional Normal Consistency (UNC) as our method produces a complete head and we are only interested in evaluating the facial region.
Despite comparing to a specialized face geometry model with fewer degrees of freedom, our results are on par, see \cref{tab:fitting_face}. While ImFace only reconstructs the face geometry, PhoMoH additionally accurately reconstructs the color and is able to realistically complete the full head. See \cref{fig:fitting_imface} for a side-by-side comparison.

In addition to ImFace, we report results for the mesh-based Basel Face Model (BFM) \cite{gerig2018morphable}. 
However, the BFM is not very diverse, especially in its appearance space as \eg evident from the second row in \cref{fig:fitting_imface}.
Thus, it is not well suited for a meaningful comparison on our test data and we report results only for completeness.

\subsection{Landmark Detection}
\label{sec:lm_dec}

\begin{figure}
    \centering
    \includegraphics[width=\linewidth]{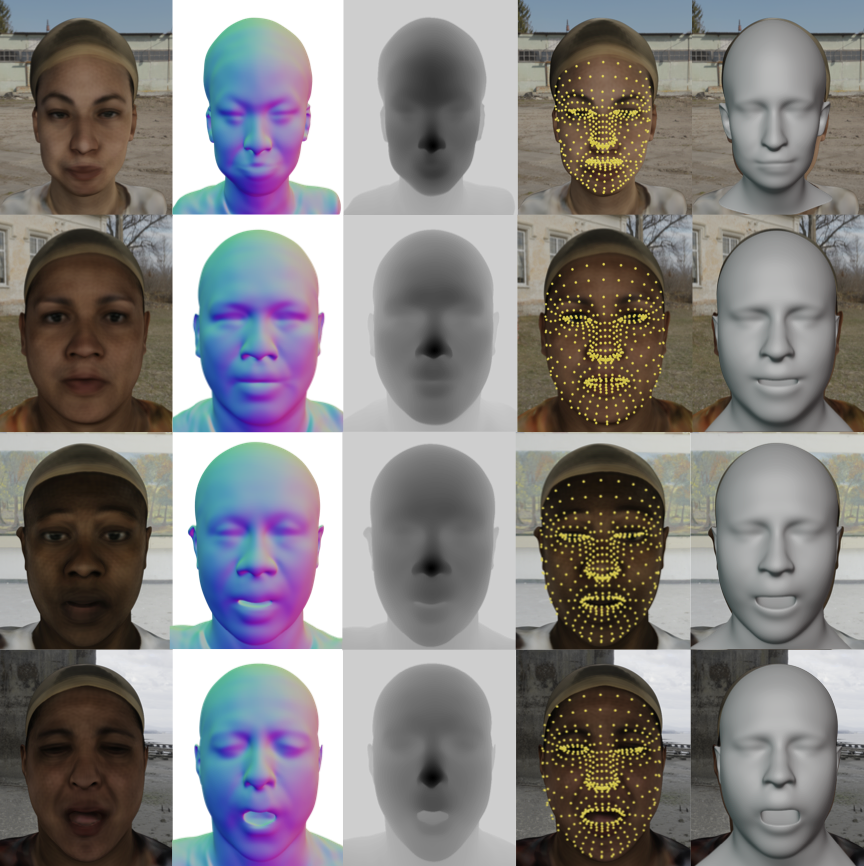}
    \caption{We can obtain various annotations for head samples from our model, making it a useful tool for dataset generation. Here we show rendered image, surface normals, depth, 3D landmarks, and the GHUM mesh.}
    \label{fig:training_data}

\end{figure}

\begin{figure}
    \centering
    \includegraphics[width=\linewidth]{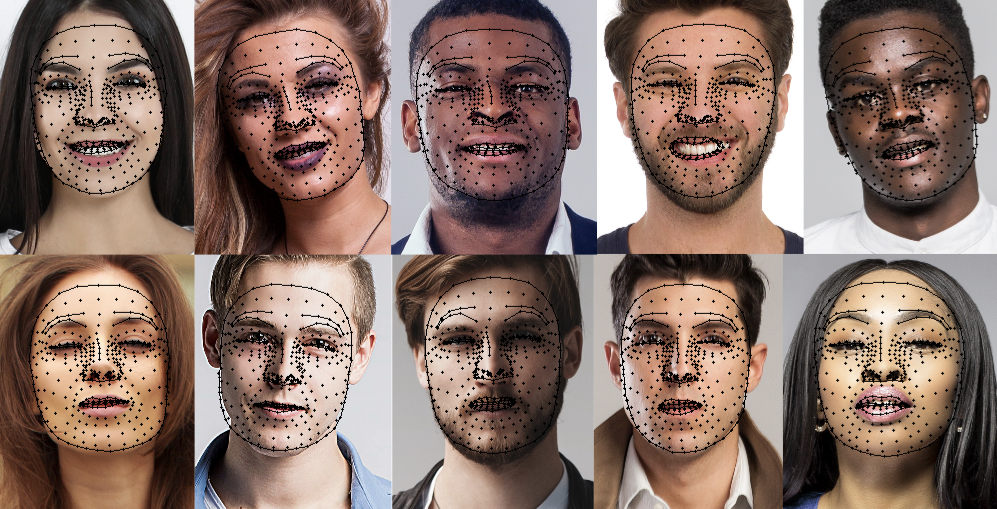}
    \caption{Qualitative results of the 2D landmark detection model trained using heads sampled from our model.}
    \label{fig:lm_detector}
\end{figure}

The nature of our model also allows its usage for dataset generation.
A number of annotations can be obtained automatically,
as we show in \cref{fig:training_data}. 
This often not the case for related image generation models, implicit representations, or neural radiance fields.
Other annotations can be transferred between instances, as we have shown in the main paper.
We demonstrate the usefulness of our model for dataset generation as follows.
Our proposed PhoMoH model allows us to associate each sampled head with a GHUM mesh approximating its geometry. We use this property to generate a synthetic dataset of pairs of rendered images together with $468$ 2D landmark positions, collected from the corresponding GHUM meshes and projected into the image plane. We generate 20k samples for training and 3k samples for validation, sampling from all the disentagled factors of our model $\featureset = (\pose,\shape,\expression,\id,\appid)$. Additionally, we sample a random camera rotation and position from which to render over a random HDRI background \cite{polyhaven}. We retrain the architecture of FaceMesh~\cite{grishchenko2020attention} using our synthetic dataset as sole supervision source. In \cref{fig:lm_detector}, we show qualitative examples demonstrating that our head samples allow to train detectors that generalize well to real images. 

{\small
\bibliographystyle{ieeenat_fullname}
\bibliography{egbib}
}

\end{document}